%% file: main.tex
\documentclass[letterpaper, 10 pt, journal, twoside]{IEEEtran} 

\IEEEoverridecommandlockouts                              % This command is only needed if 
\title{More Than a Feeling: Learning to Grasp and Regrasp using Vision and Touch}
\author{Roberto~Calandra$^{1}$, Andrew~Owens$^{1}$, Dinesh Jayaraman$^{1}$, Justin~Lin$^{1}$, Wenzhen~Yuan$^{2}$,\\ Jitendra~Malik$^{1}$, Edward~H.~Adelson$^{2}$, and Sergey~Levine$^{1}$%
\thanks{Manuscript received: February 24, 2018; Revised May 17, 2018; Accepted June 17, 2018.}%Use only for final RAL version
\thanks{This paper was recommended for publication by Editor Tamim Asfour upon evaluation of the Associate Editor and Reviewers' comments. 
This work was supported by Berkeley DeepDrive, NVIDIA, Amazon, the Toyota Research Institute, and the MIT Lincoln Labs.
}
\thanks{$^{1}$Roberto Calandra, Andrew Owen, Dinesh Jayaraman, Justin Lin, Jitendra Malik and Sergey Levine are with the Department of Electrical Engineering and Computer Sciences, University of California, Berkeley, USA
        {\tt\small \{roberto.calandra, owens, dineshjayaraman, justinlin98, malik\}@berkeley.edu, svlevine@eecs.berkeley.edu}}%
\thanks{$^{2}$Wenzhen Yuan and Edward H. Adelson are with the Massachusetts Institute of Technology, USA
	{\tt\small yuan\_wz@mit.edu, adelson@csail.mit.edu}}%
\thanks{Digital Object Identifier (DOI): 10.1109/LRA.2018.2852779}
}

\input{preamble}

\usepackage{diagbox}
\usepackage{array}
\usepackage{picinpar}
\newcolumntype{P}[1]{>{\centering\arraybackslash}p{#1}}

\newcommand{\citep}[1]{\cite{#1}}

\newcommand{\approxndatagraspingreal}[0]{{6,000}}
\newcommand{\exactndatagraspingaugmented}[0]{{18,070}}
\newcommand{\exactndatagraspingreal}[0]{{6,450}}
\newcommand{\forcegripper}[0]{F}

\newcommand{\gelsight}[0]{GelSight}
\newcommand{\nTrainingObj}[0]{65}
\newcommand{\nTestObj}[0]{22}

\newcommand{\model}[0]{f}

\begin{document}

\maketitle

%%%%%%%%%%%%%%%%%%%%%%%%%%%%%%%%%%%%%%%%%%%%%%%%%%%%%%%%%%%%%%%%%%%%%%%%%%%%%%%%

\begin{abstract}
	\input{0_abstract}
\end{abstract}

%%%%%%%%%%%%%%%%%%%%%%%%%%%%%%%%%%%%%%%%%%%%%%%%%%%%%%%%%%%%%%%%%%%%%%%%%%%%%%%%

\section{Introduction}

	\input{1_introduction}

%%%%%%%%%%%%%%%%%%%%%%%%%%%%%%%%%%%%%%%%%%%%%%%%%%%%%%%%%%%%%%%%%%%%%%%%%%%%%%%%

\section{Related Work}
\label{sec:related}

	\input{2_related}

%%%%%%%%%%%%%%%%%%%%%%%%%%%%%%%%%%%%%%%%%%%%%%%%%%%%%%%%%%%%%%%%%%%%%%%%%%%%%%%%
	
\section{Hardware Setup}
\label{sec:hardware}

	\input{2_hardware}

%%%%%%%%%%%%%%%%%%%%%%%%%%%%%%%%%%%%%%%%%%%%%%%%%%%%%%%%%%%%%%%%%%%%%%%%%%%%%%%%

\section{Deep Visuo-Tactile Models for Grasping}
\label{sec:approach}

\input{3_approach}

%%%%%%%%%%%%%%%%%%%%%%%%%%%%%%%%%%%%%%%%%%%%%%%%%%%%%%%%%%%%%%%%%%%%%%%%%%%%%%%%
	
\section{Data Collection}
\label{sec:setting}

\input{4_datacollection}

%%%%%%%%%%%%%%%%%%%%%%%%%%%%%%%%%%%%%%%%%%%%%%%%%%%%%%%%%%%%%%%%%%%%%%%%%%%%%%%%

\section{Experimental Results}
\label{sec:results}

	\input{5_results}

%%%%%%%%%%%%%%%%%%%%%%%%%%%%%%%%%%%%%%%%%%%%%%%%%%%%%%%%%%%%%%%%%%%%%%%%%%%%%%%%

\section{Conclusions}
\label{sec:conclusion}

	\input{6_conclusion}

%%%%%%%%%%%%%%%%%%%%%%%%%%%%%%%%%%%%%%%%%%%%%%%%%%%%%%%%%%%%%%%%%%%%%%%%%%%%%%%%


%%%%%%%%%%%%%%%%%%%%%%%%%%%%%%%%%%%%%%%%%%%%%%%%%%%%%%%%%%%%%%%%%%%%%%%%%%%%%%%%

\bibliographystyle{IEEEtran}
\bibliography{paper-tactile2}  % .bib

\end{document}

%% file: preamble.tex
\usepackage[english]{babel}

\usepackage{graphicx}
\usepackage{animate}
\usepackage{epsfig} 				% for postscript graphics files
\usepackage{epstopdf}

\usepackage{listings}
\usepackage{color}
\usepackage{nameref}
\usepackage{hyperref}
\usepackage[colorinlistoftodos]{todonotes}
\usepackage{amsmath}	 			% assumes amsmath package installed
\usepackage{amssymb}  				% assumes amsmath package installed

\usepackage{dsfont}			%for the symbol 'Real Numbers'
\usepackage{mathtools}

\usepackage{epigraph}
\usepackage{lscape}
\usepackage[]{nomencl}				% nomenclatures
\usepackage[ruled,vlined,linesnumbered]{algorithm2e}
\usepackage{multicol}
\usepackage{multirow}
\usepackage{etoolbox}
\usepackage{siunitx}

\usepackage{caption}
\usepackage{subcaption}

\usepackage{wrapfig}
\usepackage{multimedia}

\usepackage{units}
\usepackage{flushend}

\usepackage{floatflt}

\usepackage{url}
\usepackage[absolute,overlay]{textpos}

\usepackage{bibentry}

\usepackage{tikz}
\usepgflibrary{arrows}	% for more options on arrows

\usepackage{changes}

\usepackage{float}
\usepackage[utf8]{inputenc}
\usepackage[english]{babel}
\usepackage{graphics}
\usepackage{animate}

\usepackage{listings}

\usepackage[nostamp]{draftwatermark}

\usepackage{extarrows}		%for text above eq

\allowdisplaybreaks[1]

\newcommand{\playvideo}[1]{\href{run:#1}{\includegraphics[scale=0.12]{\RCPath fig/empty}}}

\usepackage{lipsum}
\usepackage{framed}

\input{commands.tex}

%% file: commands.tex
\newcommand{\ie}[0]{i.e.}
\newcommand{\eg}[0]{e.g.}

\newcommand{\email}[1]{\href{mailto:#1}{\nolinkurl{#1}}}

\newcommand{\link}[1]{\colora{\url{#1}}}

\renewcommand{\sec}[1]{Sec.~\ref{#1}}
\newcommand{\fig}[1]{Fig.~\ref{#1}}
\newcommand{\eq}[1]{Eqn.~\eqref{#1}}

\newcommand{\tab}[1]{Tab.~\ref{#1}}

\definecolor{matlab1}{rgb}{0,0,1}
\definecolor{matlab2}{rgb}{0,0.5,0}
\definecolor{matlab3}{rgb}{1,0,0}
\definecolor{matlab4}{rgb}{0,0.75,0.75}
\definecolor{matlab5}{rgb}{0.75,0,0.75}
\definecolor{matlab6}{rgb}{0.75,0.75,0}
\definecolor{matlab7}{rgb}{0.25,0.25,0.25}

\definecolor{darkgreen}{rgb}{0,0.5,0}		%Olivegreen?
\definecolor{purple}{rgb}{0.75,0,0.75}
\definecolor{pink}{rgb}{1,0.4,0.6}

\newcommand{\capitalize}[1]{\expandafter\MakeUppercase\expandafter{#1}}

\newcommand{\colora}[1]{{\usebeamercolor[fg]{framesubtitle}#1}}

\makeatletter
\newcommand*{\compress}{\@minipagetrue}
\makeatother

\renewcommand{\vec}[1]{\boldsymbol{#1}}				% Vector
\newcommand{\maximize}[0]{\text{arg max}} 			% Maximize

\newcommand{\q}{\vec{q}}					% Position
\ifdef{\dq}{\renewcommand{\dq}{\dot{\q}}}{\newcommand{\dq}{\dot{\q}}}
\newcommand{\MDPState}{\vec{s}} 				% State
\newcommand{\MDPAction}{\vec{a}}				% Action
\newcommand{\MDPTime}{t}					% Time

\usepackage{array}
\makeatletter
\newcommand{\thickhline}{%
    \noalign {\ifnum 0=`}\fi \hrule height 1pt
    \futurelet \reserved@a \@xhline
}
\newcolumntype{"}{@{\hskip\tabcolsep\vrule width 1pt\hskip\tabcolsep}}
\makeatother

%% file: 0_abstract.tex
For humans, the process of grasping an object relies heavily on rich tactile feedback.
Most recent robotic grasping work, however, has been based only on visual input, and thus cannot easily benefit from feedback after initiating contact. 
In this paper, we investigate how a robot can learn to use tactile information to iteratively and efficiently adjust its grasp. 
To this end, we propose an end-to-end action-conditional model that learns regrasping policies from raw visuo-tactile data. 
This model -- a deep, multimodal convolutional network -- predicts the outcome of a candidate grasp adjustment, and then executes a grasp by iteratively selecting the most promising actions.
Our approach requires neither calibration of the tactile sensors, nor any analytical modeling of contact forces, thus reducing the engineering effort required to obtain efficient grasping policies.
We train our model with data from about \exactndatagraspingreal{} grasping trials on a two-finger gripper equipped with \gelsight{} high-resolution tactile sensors on each finger. 
Across extensive experiments, our approach outperforms a variety of baselines at 
(i) estimating grasp adjustment outcomes, 
(ii) selecting efficient grasp adjustments for quick grasping, and 
(iii) reducing the amount of force applied at the fingers, while maintaining competitive performance. 
Finally, we study the choices made by our model and show that it has successfully acquired useful and interpretable grasping behaviors.

%% file: 1_introduction.tex
\IEEEPARstart{G}{rasping} is a deeply interactive task: we initiate contact by reaching our fingers toward an object, adjust the placement of our fingers, and balance contact forces as we lift.
During this process, the feedback provided by the sense of touch is paramount, as demonstrated by human experiments~\citep{Johansson2009}.
Nonetheless, incorporating touch sensing into robotic grasping has thus far proved challenging, due to hardware limitations (\eg, sensor sensitivity and cost) and the difficulty of integrating tactile inputs into standard control schemes. 
Consequently, the predominant input modalities currently used in the robotic grasping literature are vision and depth.

\begin{wrapfigure}{r}{0.53\linewidth}
	\vspace{-10pt}
	\centering
	\includegraphics[width=0.99\linewidth]{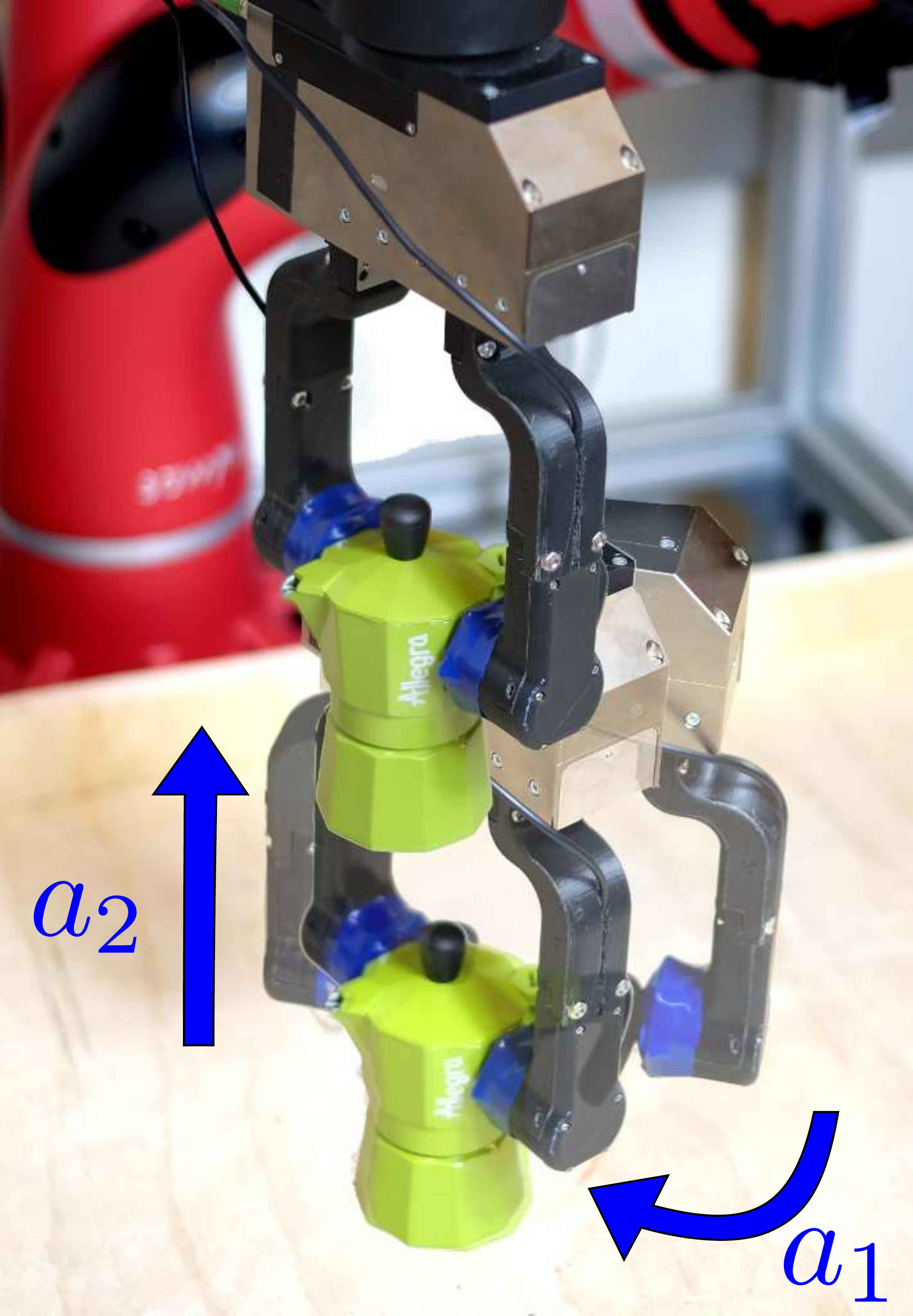}
	\caption{We propose an action-conditional model that iteratively adjusts a robot's grasp based on raw visuo-tactile inputs.}
	\label{fig:setting}
	%     \vspace{-10pt}
\end{wrapfigure}
However, vision does not easily permit the measurement of and reaction to ongoing contact forces, thus significantly hindering the potential benefits of interaction.
As a result, vision-based grasping approaches have largely relied on selecting a grasp configuration (location, orientation, and forces) in advance, before making contact with the object.

In the quest for interactive grasping, we study how tactile sensing can be integrated into a grasping system that can probe an object and then reactively adjust its grasp to achieve the highest chance of success.
Our method is based on learning an action-conditioned grasping model, trained end-to-end in a self-supervised manner by using a robot to autonomously collect grasp attempts.
In contrast to prior self-supervised grasping work~\citep{Pinto2016,Levine2016}, however, our model incorporates rich touch sensing from a pair of \gelsight{} sensors (see \fig{fig:setting}).
Incorporating tactile sensing into action-conditional models, however, is not straightforward. 
The robot only receives tactile input intermittently, when its fingers are in contact with the object and, since each regrasp attempt can disturb the object position and pose, the scene changes with each interaction.
In contrast, grasping methods that use vision typically do not interact repeatedly with the object, but simply drive the arm toward a chosen grasp pose and then attempt a single grasp.
 
Our contributions are as follows: 
(1) we introduce a new multi-modal action-conditional model for grasping using vision and touch;
(2) we show that our model is effective at grasping novel objects, in comparison to unconditional models and vision-only variations; 
(3) we analyze the learned grasping policy and show that it produces interpretable and useful grasping behaviors;
(4) we demonstrate that our model permits explicit constraints on contact forces, allowing us to command the robot to ``gently" grasp an object with significantly reduced force. 
Since it incorporates raw visuo-tactile inputs, our approach requires neither calibration of the tactile sensors, nor any analytical modeling of contact forces, hence significantly reducing the engineering effort required to obtain efficient grasping policies. 

\newcommand{\heighttac}{2.6cm}
\begin{figure*}[t]
	\centering
	\includegraphics[height=\heighttac]{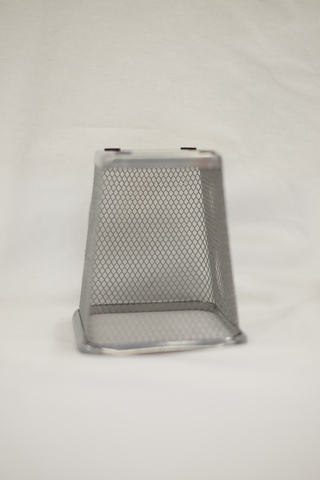}%
	\includegraphics[height=\heighttac]{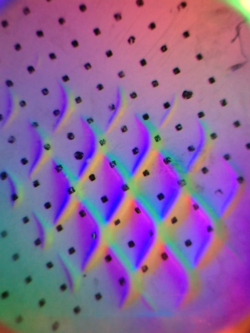} \hfill
	\includegraphics[height=\heighttac]{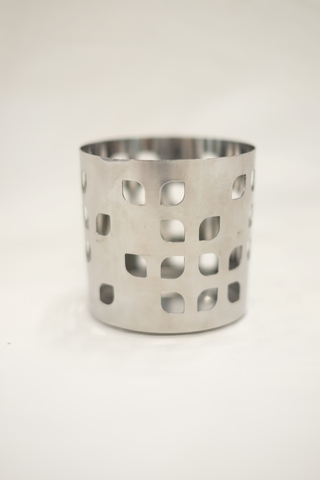}%
	\includegraphics[height=\heighttac]{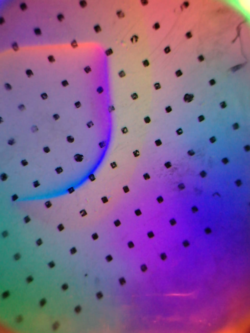} \hfill
	\includegraphics[height=\heighttac]{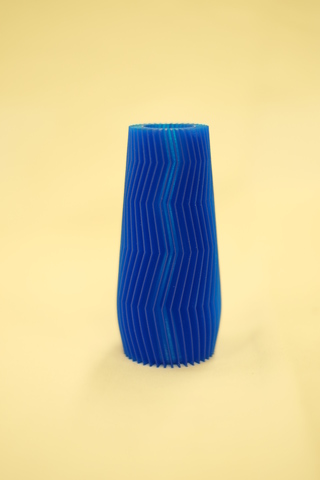}%
	\includegraphics[height=\heighttac]{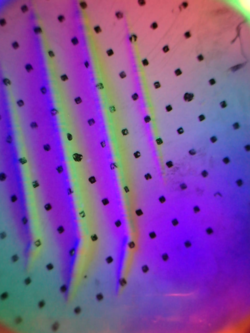} \hfill
	\includegraphics[height=\heighttac]{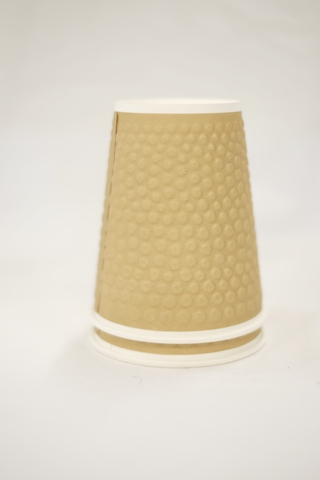}%
	\includegraphics[height=\heighttac]{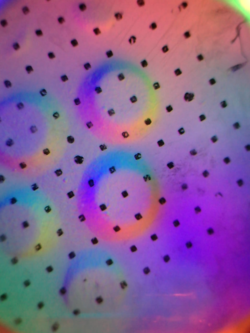}
	\\ \vspace{3pt}
	\includegraphics[height=\heighttac]{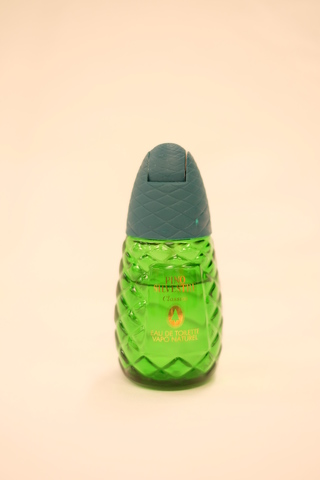}%
	\includegraphics[height=\heighttac]{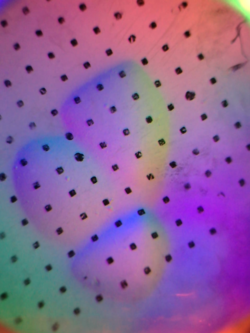} \hfill
	\includegraphics[height=\heighttac]{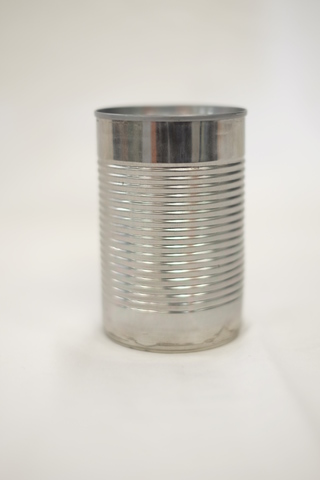}%
	\includegraphics[height=\heighttac]{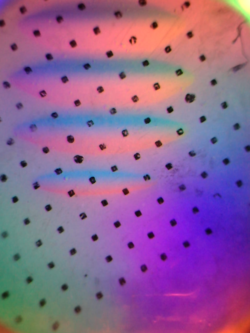} \hfill
	\includegraphics[height=\heighttac]{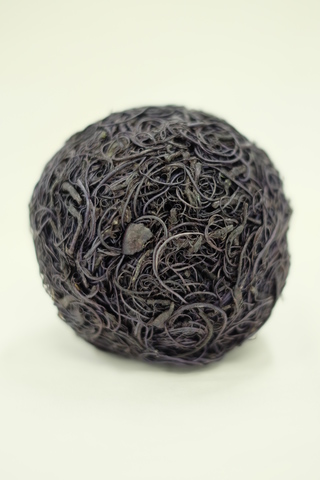}%
	\includegraphics[height=\heighttac]{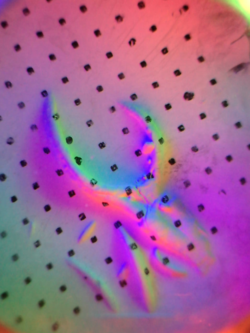} \hfill
	\includegraphics[height=\heighttac]{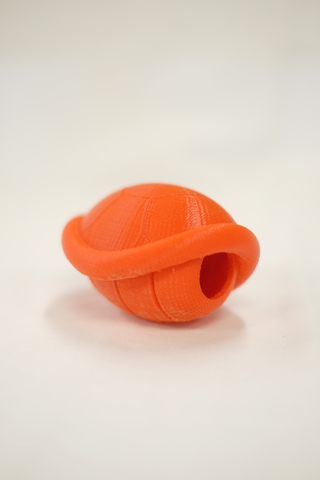}%
	\includegraphics[height=\heighttac]{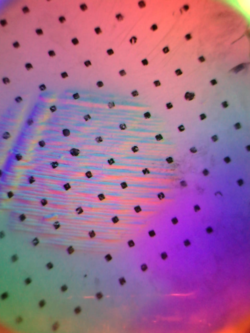}
	\caption{Examples of raw tactile data collected by one of the \gelsight{}s (\textit{right}) for different training objects (\textit{left}).}
	\label{fig:tactile_data}
%      \vspace{-10pt}
\end{figure*}
%

%% file: 2_related.tex
\subsection{Learning to Grasp}
	A significant body of work in robotics has studied analytic grasping models, which use known or estimated models of object geometry, environments, and robot grippers, and which typically make use of manually defined grasping metrics~\citep{Shimoga1996,Goldfeder2011,Rodriguez2012}. 
	While these methods provide considerable insight into the physical interactions in grasping, their actual performance depends on how well the real-world system fits the assumptions of the analytic model. 
	Model misspecification and unmodeled factors can substantially reduce their effectiveness.
	As an alternative, data-driven approaches have sought to predict grasp outcomes from human supervision~\citep{Kamon1996,Lenz2015}, simulation~\citep{Kappler2015,Johns2016,Mahler2016}, or autonomous robotic data collection~\citep{Pinto2016,Levine2016}, typically using visual or depth observations.
	Among these works, the most related to ours is~\citep{Levine2016}, which also proposes to use an action-conditional model.
	However, these prior works (with a few exceptions that we discuss below) do not consider tactile sensing, focusing instead on vision and 3D geometry, which afford a limited ability to reason about contact forces, pressures, and compliance. 
	Critically, most of these methods rely on selecting grasp configurations in advance, before ever coming into contact with the target object. 
	In contrast, we show that it is possible to exploit rich tactile feedback \emph{after contact} to iteratively adjust and improve robotic grasps.
	For an overview of learning for robot grasping, we refer the reader to~\citep{Bohg2014}. 

\subsection{Tactile Sensors in Grasping}

	A variety of tactile sensors have been developed~\citep{Yousef2011}, mainly measuring force and torque, or the pressure distribution over the sensor.
	Multiple works \citep{Bekiroglu2011,Schill2012,Dang2014,Cockbum2017,Calandra2017} suggested the use of tactile sensors to estimate grasp stability. 
	While these works estimate the stability of an ongoing grasp, we focus instead on \emph{selecting} grasp adjustments to produce a stable new grasp.
	\citep{Li2014a} incorporated tactile readings into dynamics models of objects for a dexterous hand, thereby adapting the grasp.
	Works such as \citep{Bicchi1988,Romano2011,Veiga2015} extracted features from tactile signals to detect/predict slip, so as to adaptively adjust the grasping force. 
	Researchers have also proposed robotic systems that integrate visual and tactile information for grasping using model-based methods~\citep{Allen1999,Bekiroglu2012,Jara2014,Bekiroglu2016,Guo2017}, which improved grasping performance over single-modality inputs. 
	However, these approaches require accurate models of the robot and the objects to grasp, and often also calibrated tactile sensors. 
	Along similar lines, \citep{Hyttinen2017} proposed a regrasping policy based on tactile sensing (without visual input) and a learned stability metric, which uses a heuristic transition function to predict future tactile readings.
	Our approach does not require any prior model or transition function, as it learns entirely end-to-end from raw inputs.

	Closer to our approach are~\citep{Chebotar2016,Chebotar2016a}, which proposed to learn regrasping using tactile sensors.
	In contrast to our approach, \citep{Chebotar2016a} directly optimizes a policy. 
	Optimizing a policy requires the data collection to be on-policy and to be intertwined with the policy update; our approach does not directly optimize a policy, but learns an action-conditioned model. 
	As a result our approach can use any data collected.
	Additionally, by using an action-conditioned model, we can change the objective of the policy at evaluation time (as in the case of reducing the grasping force demonstrated in \sec{sec:results:minimumForce}), while changing the objective for a policy learning method would require re-training the policy, and thus require repeating the data collection process.
	Another difference with these works is that, in~\citep{Chebotar2016,Chebotar2016a}, the features used from the tactile sensors are manually designed by applying PCA and extracting the first 5 principal components. 
	Our approach, although using substantially higher resolution tactile inputs, does not require any manual engineering of features.
	Finally, our experiments consider a substantially wider range of objects than demonstrated by~\citep{Chebotar2016a}, with \nTrainingObj{} training objects, and a detailed evaluation on \nTestObj{} previously unseen test objects.

	% Our previous paper
	Closely related is also our previous work~\citep{Calandra2017}, where we proposed a visuo-tactile model from raw inputs for classifying grasp outcomes. 
	The main difference to the present work is that~\citep{Calandra2017} does not make use of the learned visuo-tactile model to actively select the next grasp to perform, but simply to evaluate the stability of an ongoing grasp.  
	For grasp selection, this method executes random grasps iteratively until it arrives at a grasp that is stable according to the learned model. 
	While this allows for evaluation of the correlation between touch sensing and grasp outcome, it does not by itself provide a practical method for grasp selection: in our experiments, we found that this prior approach could require as many as 50 random regrasp attempts to yield a stable grasp. 
	Furthermore, by including the grasping force as part of the action, our approach allows for the grasping force to be modulated during the evaluation to achieve secondary objectives, such as minimum-force grasps.
	
	Concurrently to our work, \citep{Hogan2018} also proposed a tactile regrasping method based on the GelSight sensor. This method simulates transformations to tactile readings based on rigid body dynamics, while our approach is entirely data-driven and self-supervised, which means that we do not require assumptions about dynamics or environment structure. 
	An in-depth exploration of the tradeoffs between data-driven and analytic approaches would an interesting future topic of study.
	Another concurrent work~\citep{Murali2018}, explores grasping with a 3-axis force sensor, but reports comparatively low success rates, focusing instead on tactile localization without vision. Our method uses rich touch sensing that is aware of texture and surface shape, simultaneously incorporates multiple modalities, and can flexibly accommodate additional constraints, such as minimum-force grasps.
	
	The main contribution of this paper is a practical approach that exploits visual and tactile sensing to grasp successfully and efficiently \ie{}, with as few regrasps as possible.
	We do so by building predictive models that can predict the grasp outcome of a given action. 
	Our experiments demonstrate that our action-conditioned predictive model substantially outperforms the results that can be obtained via grasp classification, illustrating the value of closed-loop regrasping. 
	Finally, we demonstrate that our action-conditioned model can be used to optimize for gentler grasps, enabling the robot to determine grasps that can pick up an object with minimal force (hence avoiding damage to fragile objects).
	To the best of our knowledge, our work is the first to propose an action-conditioned model for learning to grasp from raw visuo-tactile inputs.

%% file: 2_hardware.tex
% Hardware setting
In our experiments we used a hardware configuration consisting of a 7-DoF Sawyer arm, a Weiss WSG-50 parallel gripper, and two \gelsight{} sensors~\citep{Yuan2017b}, one for each finger.
Each \gelsight{} sensor provides raw pixel measurements at a resolution of 1280x960 at \SI{30}{\hertz} over an area of \SI{24x18}{\milli\meter}. 
Additionally, a Microsoft Kinect2 sensor was mounted in front of the robot to provide visual data.
%
% Gelsights 
The \gelsight{} sensor is an optical tactile sensor that measures high-resolution topography of the contact surface~\citep{Johnson2009,Yuan2017b}. 
The surface of the sensor is a soft elastomer painted with a reflective membrane, which deforms to the shape of the object upon contact. 
Underneath this elastomer is a camera (an ordinary webcam) that views the deformed gel. 
The gel is illuminated by colored lights, which light the gel from different directions.
Additional visual cues of contacts are provided by the deformation of the grid of markers painted on the sensor surface, which can be used to compute the shear force and slip information~\citep{Yuan2015}.
One valuable property of the \gelsight{} sensor is that the sensory data is provided on a regular 2D grid image format, hence we can use convolutional neural network (CNN) architectures initially designed for visual processing to process readings from the tactile sensor.
Previous work on material property estimation with \gelsight{}~\citep{Yuan2017,Yuan2017a} has successfully applied CNNs pretrained from natural image data.
Examples of raw tactile data from the \gelsight{} are shown in \fig{fig:tactile_data}.

%% file: 3_approach.tex
%\subsection{Problem setup}
\label{sec:notation}
We formalize grasping as a Markov decision process~(MDP) where we greedily select the gripper actions that maximize the probability of successfully grasping an object. 
To address this, we solve the following prediction problem: given the robot's current visuo-tactile observations~$\MDPState_\MDPTime$ at time $\MDPTime$, and an action~$\MDPAction$, we predict the probability that, after applying the action, the gripper will be in a configuration that leads to a successful grasp at time $\MDPTime+1$.
In \sec{sec:regrasp}, we describe how we use this prediction model to select optimal grasping actions.

\begin{figure*}[t]
	\centering
	\includegraphics[width=0.96\linewidth]{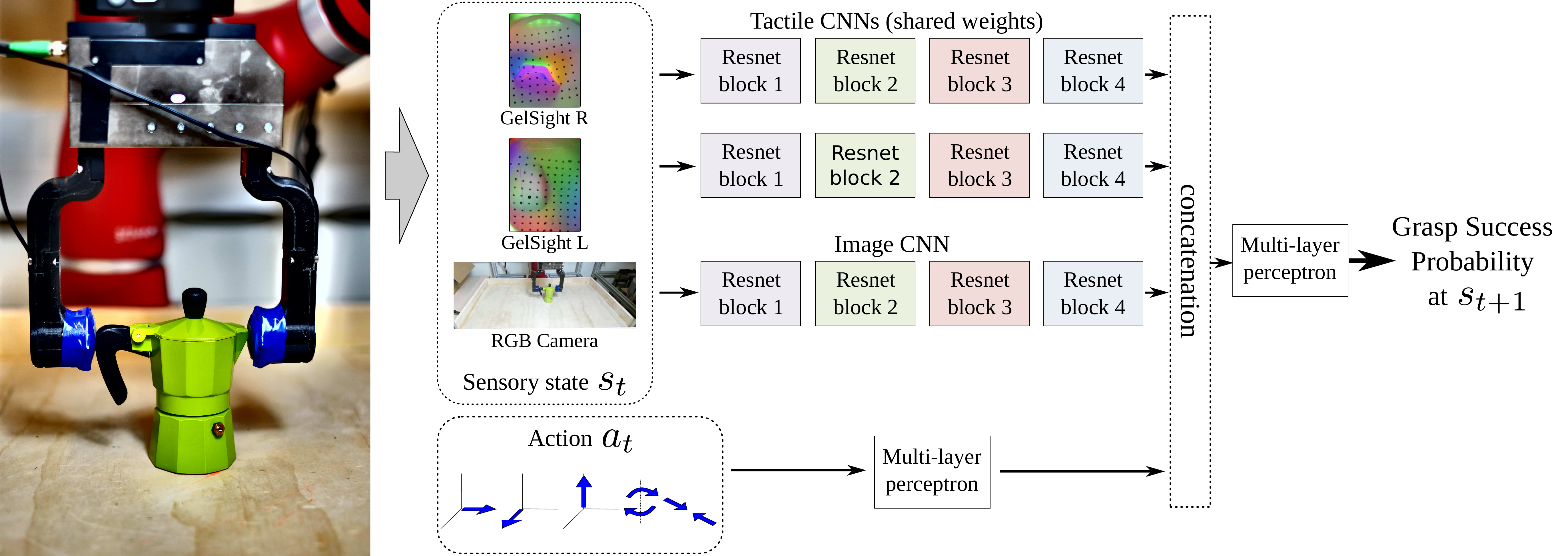}
	\caption{Action-conditioned visuo-tactile model network architecture.}
	\label{fig:model}
% 	\vspace{-10pt}
\end{figure*}

Raw visuo-tactile observations $\MDPState$ are acquired from tactile sensors and the RGB camera, as shown in \fig{fig:model}. 
Each action~$\MDPAction$ directs the gripper to a new pose relative to its current pose. 
For example, an action $\MDPAction$ might consist of moving the gripper to the left by $\SI{2}{\centi\meter}$, and rotating it by $\ang{15}$.
More concretely, let $o(\MDPState_\MDPTime,\MDPAction)\in\{0,1\}$ be the binary grasp outcome at time $\MDPTime+1$ resulting from executing action $\MDPAction$ from grasp state $\MDPState_\MDPTime$: if $o(\MDPState, \MDPAction)$ is 1, the grasp is successful. At evaluation time, these outcome labels $o(\MDPState_\MDPTime,\MDPAction_\MDPTime)$ are unknown and the robot must estimate them. 
At training time, the robot performs random trials as described in \sec{sec:setting} to collect state-action-outcome tuples $(\MDPState_i,\MDPAction_i,o_i) \in X$, which we will use to train an action-conditional model that can be used for selecting actions.

\subsection{End-to-End Outcome Prediction}
\label{sec:approach:end-to-end-math}

	We would like to learn a function $\model(\MDPState,\MDPAction)$ that directly predicts the success probability for a future grasp, given observations from the current grasp $\MDPState$ and a candidate action~$\MDPAction$. 
	We parametrize $\model$ as a deep neural network, whose architecture is shown in \fig{fig:model}. 
	There are multiple design choices when designing deep models for multi-modal inputs~\citep{Ngiam2011}.
	In our experiments, we decided to employ a network processing the state~$\MDPState$, consisting of raw RGB inputs from the frontal camera and the two GelSight tactile sensors, in three deep stacks of convolutional layers. 
	Additionally, the action~$\MDPAction$ is processed in a two-layer, fully-connected stack (a multi-layer perceptron). 
	We then use a late fusion approach to combine information from these modalities: the feature vectors produced by these four stacks are concatenated, and fed to a two-layer fully-connected network that produces the probability, $\model(\MDPState,\MDPAction)$, that the input action from the current state results in a successful grasp at the next step.
	We train the network~$\model$ on the training dataset $X$ to minimize the loss
	$L_{dir}(f,X) = \sum_{(\MDPState,\MDPAction,o)\in X} \mathcal{L}(\model(\MDPState,\MDPAction), o)$
	where $\mathcal{L}$ is the cross-entropy loss.
	As input for the tactile CNNs, we rescale the original GelSight RGB images to $256 \times 256$, and subsequently (for data augmentation) sample random $224 \times 224$ crops. 
	This kind of image resolution is standard for CNN-based object recognition in computer vision, though it is substantially lower than the native resolution of the GelSight. 
	Although we did not investigate the effect of image resolution on performance, this is an interesting question for future work.

	\paragraph{Network design} 
		We process each image using a convolutional network. 
		Specifically, we use the penultimate layer of a 50-layer deep residual network~\citep{He2016}.  
		We further emphasize deformations in each \gelsight~image through background subtraction i.e., we pass the neural network the difference of the \gelsight~images before and after contact.
		The action network is a multi-layer perceptron consisting of two fully-connected layers with 1024 hidden units each.  
		This network takes as input vector representations of the action and pose. 
		The action is a 5-dimensional vector consisting of a 3D motion, in-plane rotation, and change in force.  
		Likewise, the end effector pose is a 4-dimensional vector represented by position and angle.  
		Moreover, we also provided the network with the 3D motion transformed into the gripper's coordinate system.
		To fuse these networks, we concatenate the outputs of the four input branches (camera image, two \gelsight~ images, and the action network), and then pass them through a two-layer fully-connected network that produces a grasp success probability.  
		The first layer of this fusion network contains 1024 hidden units. 
		Our model architecture is shown in \fig{fig:model}.

	\paragraph{Training}
		To speed up training, we pretrain these networks using weights from a model trained to classify objects on ImageNet~\citep{deng2009imagenet}, and we tie the weights of the two tactile networks. 
		We then jointly optimize the model with a batch size of 16 for 9,000 iterations
		(using a dataset of \exactndatagraspingaugmented~examples), lowering the learning rate by a factor of 10 after 7000 iterations.

\subsection{Regrasp Optimization}
\label{sec:regrasp}

	Once the action-conditional model~$\model$ has been learned, we use it to select the action that maximize the expected probability of success of the grasp \emph{after} performing the action
	\vspace{-10pt}
	\begin{align}
		\MDPAction_\MDPTime^* &= \maximize_{\MDPAction}\, f\left(\MDPState_{\MDPTime}, \MDPAction\right)\,.
		\label{eq:max-over-next-reward}
	\end{align}
	We perform this optimization using stochastic search: we randomly sample potential actions and predict the success probability using the learned model~$\model$, and then select the action with the highest success probability. 
	Although this optimization can be computationally expensive (in our experiments, approximately \SI{0.6}{\second} for 5000 samples), in practice we find that it performs well.

%% file: 4_datacollection.tex
% Data collection & processing
To collect the data necessary to train our model, we designed a self-supervised automated data collection process.
In each trial, depth data from the front Kinect was used to approximately identify the starting position of the object and enclose it within a cylinder.
We then set the end-effector $(x,y)$ coordinates to the position of the center of the cylinder plus a small random perturbation, and set its height to be a random value between the floor and the height of the cylinder. 
Its orientation $\phi$ was set uniformly at random.
Moreover, we randomized the gripping force~$\forcegripper$ to collect a large variety of behaviors, from firm, stable grasps, to occasional slips, to overly gentle grasps that fail more often.
After moving to the chosen position and orientation, and closing the gripper with the desired gripping force, the gripper attempt to lift the object and wait in the air for \SI{4}{\second}.
If the object was still in the gripper at the end of this time, the robot would place the object back at a randomized position, and a new trial would start.

The labels for this data (\ie, whether the grasp was successful) were also automatically generated using deep neural network classifiers (running two instances, one for each finger) trained to detect contacts using the raw \gelsight{} images observed\footnote{This model was initially trained using manually collected data, and iteratively fine-tuned in a self-supervised manner using the very same automatically collected, but manually labeled, data.}.
We performed additional manual labeling on a small set of samples for which the automatic classification was borderline ambiguous (\eg, if both sensor were not confident of the presence of contacts after lifting), or in the rare cases when a visual inspection would indicate a wrong label.
Overall, we collected \exactndatagraspingreal{} grasping trials from over \nTrainingObj{} training objects.

As the gripper moves from one position to another, the locations that it moves to along the way can provide additional data points for training.  
We use this idea to augment the dataset with additional examples. 
When the robot is gripping an object, we create a state-action pair with zero translation or rotation, corresponding to the action of the robot keeping the gripper in the same position (a useful possible action for regrasping).  
Similarly, we create a state-action pair at the moment that the robot has released the gripper but has not yet moved.  
In this case, the action is the same as when the gripper is in contact with the object. 
After this augmentation, our dataset contains \exactndatagraspingaugmented~examples.

During the data collection and experimental evaluation, we replaced the gels of the two \gelsight{} sensors multiple times due to wear and tear.
Each gel is unique, and as a result produces slightly different inputs (\eg, grid of markers might not be evenly aligned).
Moreover, with the progressive wear of the surface a single gel, the images can significantly change over time.
In our experiments we noticed how, initially, replacing the gel would degrade the performance of the learned models.
However, after collecting data with a few different gels, changing the gels did not seem to significantly affect performance anymore, hence suggesting that the model learned features that are reasonably invariant to the specific gel being used.

%% file: 5_results.tex
% We now present the experimental results.
To validate our multi-modal grasping model, we first compare the performance of the model on the dataset we collected. 
Then, we test the model on an actual robot, and evaluate its generalization capabilities on additional (unseen) test objects.
Moreover, we analyze the learned visuo-tactile model to gain some insight into its learned behavior and features.
Finally, we demonstrate that it is possible to exploit our visuo-tactile action-conditioned model to minimize the applied forces while maintaining a high success rate.
Videos showing the robotic grasping experiments (and other material) are available online at: \url{https://sites.google.com/view/more-than-a-feeling}

\subsection{Model Evaluation}
\label{sec:results:model}

\begin{wraptable}{r}{0.65\linewidth}
% \vspace{-10pt}
% 	\begin{table}[t]
		\centering
		\caption{K-fold (K=3) cross-validation accuracy of the different models trained with \exactndatagraspingaugmented{} data points.}
		\label{tab:model}
\resizebox{\linewidth}{!}{% 
    \begin{tabular}{|l|P{2.6cm}|}
		  \hline 
		  \multirow{2}{*}{Model} & Accuracy (mean~$\pm$~std.~err.)\\
		  \hline 
		  Chance & $62.80\% \pm 0.85\%$ \\
      Vision (+ action) & $73.03\% \pm 0.24\%$ \\
      Tactile (+ action)  & $79.34\% \pm 0.66\%$ \\
      Tactile + Vision (+ action) & $\mathbf{80.28\% \pm 0.68\%}$ \\
      Tactile + Vision (no action)&  $76.43\% \pm 0.42\%$ \\
		  \hline 
		\end{tabular}
}
% \vspace{-3mm}
% \end{table}
% \vspace{-10pt}
\end{wraptable}
First, we ask: can our model successfully learn to predict future grasp success for novel objects? 
Recall that while previous works such as~\citep{Calandra2017} have shown that it is possible to predict stability of ongoing grasps from visuo-tactile inputs, we seek to evaluate the stability of \emph{future} grasps, conditional on a relative adjustment from the current grasp. 
We compare the predictive performance of a number of variations of our model, using our dataset of grasps (\sec{sec:setting}). 
For this, we use K-fold ($K=3$) cross-validation, partitioning the data by object instance. 
Does our model learn to use actions to predict future outcomes? 
This is critical, since we expect to use this model to search over possible actions during grasping on a robot. 
To test this, we evaluate the model in \fig{fig:model} without the action (``Tactile + Vision (no action)'' in \tab{tab:model}) -- an unconditional model similar to the one considered in~\citep{Calandra2017} -- which without having access to the action corresponds to computing the expectation over all the possible actions. 
We see that performance indeed drops significantly when action information is withheld, validating that the model learns to successfully evaluate the importance of different actions. 
Next, we test whether our model significantly outperforms variations where different components are ablated, such as the vision-only and tactile-only models. 
As seen in \tab{tab:model}, the full visuo-tactile model performs best -- results for future-grasp prediction that are consistent with those reported in~\citep{Calandra2017} for the task of evaluating current grasps.

\subsection{Robot Grasp Evaluation}
\label{sec:results:grasp}

	\newcommand{\heightfig}[0]{1.4cm}
	\begin{table*}[t]
		\centering
		\caption{Detailed grasping results using different policies for the "Easy" and "Hard" test objects.}
		\label{tab:objects}
		\resizebox{\linewidth}{!}{% 
			\begin{tabular}{|P{0.3cm}|P{1.9cm}|P{1.65cm}@{\hskip 2pt}P{1.65cm}@{\hskip 2pt}P{1.65cm}@{\hskip 2pt}P{1.65cm}@{\hskip 2pt}P{1.65cm}@{\hskip 2pt}P{1.65cm}@{\hskip 2pt}P{1.65cm}@{\hskip 2pt}P{1.65cm}@{\hskip 2pt}P{1.65cm}@{\hskip 2pt}P{1.65cm}@{\hskip 2pt}P{1.65cm}|P{.9cm}|}
				\hline
				\multirow{12}{*}{\rotatebox[origin=c]{90}{\large ``Easy'' set}} &\multirow{2}{*}{\diagbox[height=2.44cm,width=2.34cm]{\hspace{10pt}Methods}{\vspace{7pt}\\Objects}}  & \raisebox{6pt}{\includegraphics[height=\heightfig, angle=270]{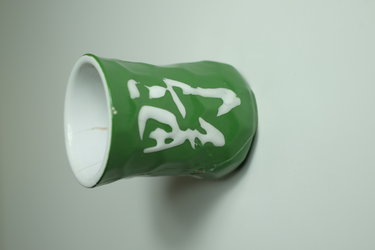}} & \raisebox{6pt}{\includegraphics[height=\heightfig, angle=270]{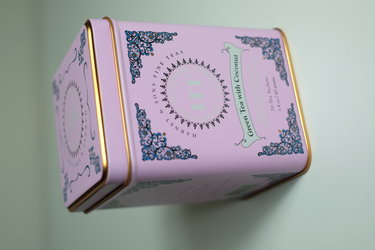}} & \raisebox{6pt}{\includegraphics[height=\heightfig, angle=270]{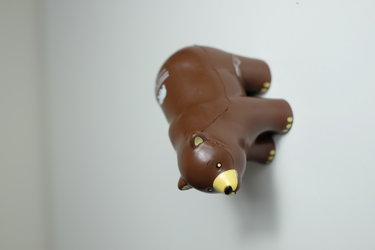}} & \raisebox{6pt}{\includegraphics[height=\heightfig, angle=270]{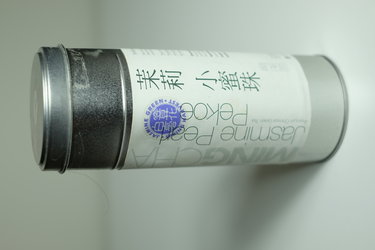}} & \raisebox{6pt}{\includegraphics[height=\heightfig, angle=270]{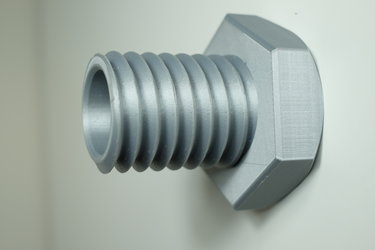}} & \raisebox{6pt}{\includegraphics[height=\heightfig, angle=270]{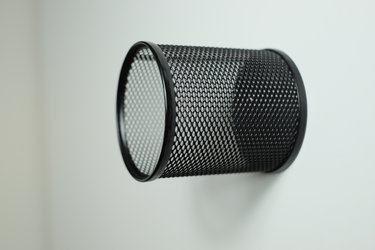}} & \raisebox{6pt}{\includegraphics[height=\heightfig, angle=270]{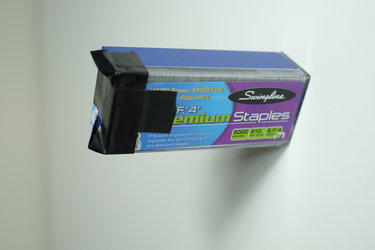}} & \raisebox{6pt}{\includegraphics[height=\heightfig, angle=270]{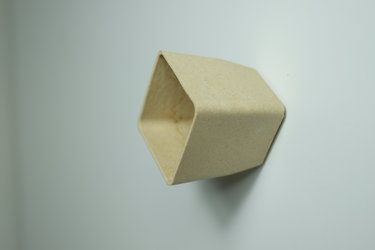}} & \raisebox{6pt}{\includegraphics[height=\heightfig, angle=270]{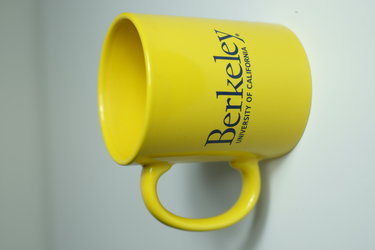}} & \raisebox{6pt}{\includegraphics[height=\heightfig, angle=270]{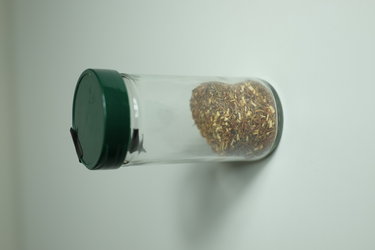}} & \raisebox{6pt}{\includegraphics[height=\heightfig, angle=270]{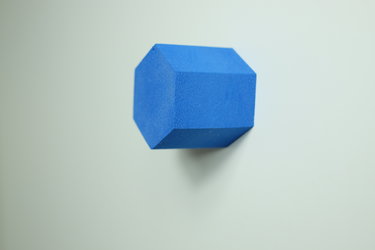}} & \vspace{20pt}Average grasp success\\ 
				& & $215\si{\gram}$ & $160\si{\gram}$ & $40\si{\gram}$ & $125\si{\gram}$ & $125\si{\gram}$ & $65\si{\gram}$ & $135\si{\gram}$ & $30\si{\gram}$ & $380\si{\gram}$ &$140\si{\gram}$ & $10\si{\gram}$ & \multicolumn{1}{c|}{}\\ \cline{3-13}
				& &  \multicolumn{11}{c|}{\% grasp success (\# success / \# trials)} & \\ \cline{2-14}
				& \multicolumn{1}{l|}{Vision only} & 76\% (38/50) & 70\% (7/10) & 60\% (6/10) & 50\% (5/10) & 50\% (5/10) & 90\% (9/10)&  40\% (4/10)& 60\% (6/10)& \textbf{90\% (9/10)} & 10\% (1/10)& \textbf{100\% (10/10)} &63.2\%\\
				& \multicolumn{1}{l|}{Tactile + Vision} & \textbf{95\% (95/100)} & \textbf{100\% (10/10)} & \textbf{100\% (10/10)} & \textbf{100\% (10/10)} & 90\% (9/10)& \textbf{100\% (10/10)}& \textbf{90\% (9/10)}& \textbf{100\% (10/10)}& 80\% (8/10) & \textbf{90\% (9/10)} &  90\% (9/10) &\textbf{94.0\%} \\ 
				& \multicolumn{1}{l|}{Cylinder fitting} & 90\% (18/20) & 90\% (18/20) & 80\% (16/20) & 55\% (11/20) &  \textbf{100\% (20/20)} & \textbf{100\% (20/20)} & \textbf{90\% (18/20)} & 75\% (15/20) & 35\% (7/20) & 20\% (4/20) & \textbf{100\% (20/20)} &75.9\%  \\ \hline\hline
				\multirow{12}{*}{\rotatebox[origin=c]{90}{\large ``Hard'' set}} & \multirow{2}{*}{\diagbox[height=2.44cm,width=2.34cm]{\hspace{10pt}Methods}{\vspace{7pt}\\Objects}}  & \raisebox{6pt}{\includegraphics[height=\heightfig, angle=270]{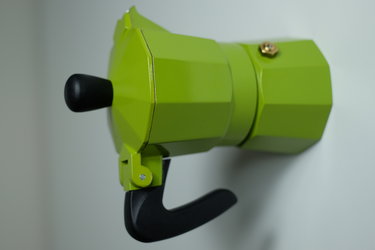}} & \raisebox{6pt}{\includegraphics[height=\heightfig, angle=270]{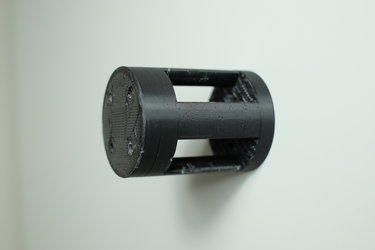}} & \raisebox{6pt}{\includegraphics[height=\heightfig, angle=270]{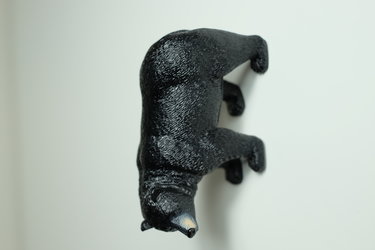}} & \raisebox{6pt}{\includegraphics[height=\heightfig, angle=270]{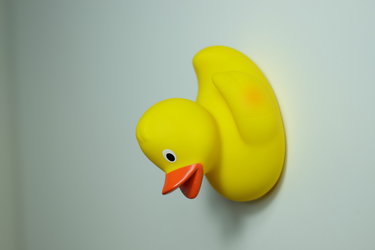}} & \raisebox{6pt}{\includegraphics[height=\heightfig, angle=270]{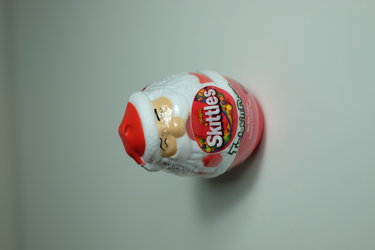}} & \raisebox{6pt}{\includegraphics[height=\heightfig, angle=270]{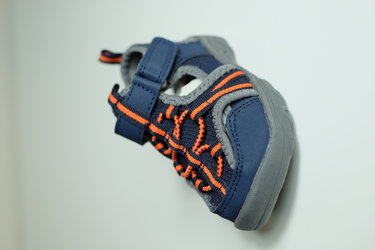}} & \raisebox{6pt}{\includegraphics[height=\heightfig, angle=270]{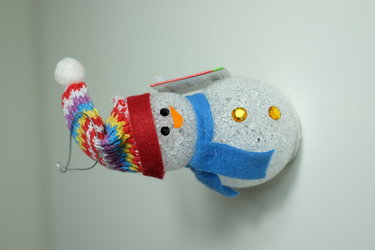}} & \raisebox{6pt}{\includegraphics[height=\heightfig, angle=270]{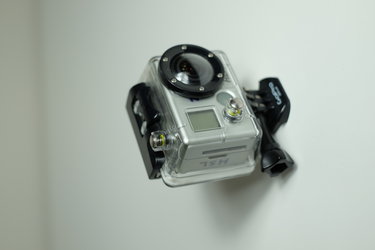}} & \raisebox{6pt}{\includegraphics[height=\heightfig, angle=270]{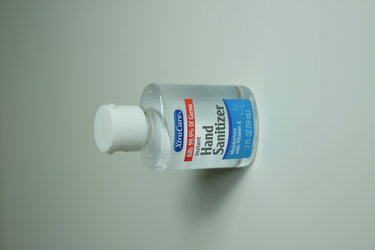}} & \raisebox{6pt}{\includegraphics[height=\heightfig, angle=270]{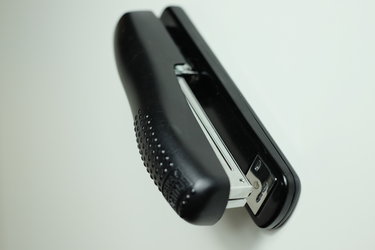}} & \raisebox{6pt}{\includegraphics[height=\heightfig, angle=270]{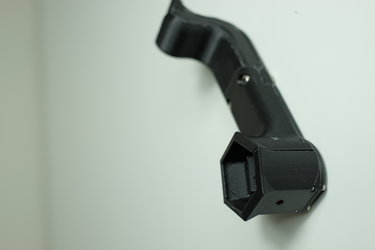}} & \vspace{20pt}Average grasp success\\ 
				& & $230\si{\gram}$ & $120\si{\gram}$ & $195\si{\gram}$ & $50\si{\gram}$ & $70\si{\gram}$ & $85\si{\gram}$ & $38\si{\gram}$ & $165\si{\gram}$ & $65\si{\gram}$ &$340\si{\gram}$ & $110\si{\gram}$ & \multicolumn{1}{c|}{}\\ \cline{3-13}
				& &  \multicolumn{11}{c|}{\% grasp success (\# success / \# trials)} & \\ \cline{2-14}
				& \multicolumn{1}{l|}{Vision only} & 60\% (6/10) & 80\%  (8/10) & 30\% (3/10) & 30\% (3/10) & 80\% (8/10) & 40\% (4/10) & 60\% (6/10) & 50\% (5/10) & 50\% (5/10) & 50\% (5/10) & 20\% (2/10) & 50\%\\
				& \multicolumn{1}{l|}{Tactile + Vision} & 80 \% (8/10) & \textbf{100\% (10/10)} & \textbf{50\% (5/10)} & 80\% (8/10) & \textbf{90\% (9/10)} & \textbf{70\% (7/10)} &  \textbf{100\% (10/10)} & 40\% (4/10) & \textbf{60\% (6/10)} & \textbf{80\% (8/10)} & 60\% (6/10) &\textbf{73.6\%}\\ 
				& \multicolumn{1}{l|}{Cylinder fitting} & \textbf{95\% (19/20)} & \textbf{100\% (20/20)} & 35\% (7/20) & \textbf{100\% (20/20)} & \textbf{90\% (18/20)} & 15\% (3/20) & 90\% (18/20) & \textbf{85\% (17/20)} & 15\% (3/20) & 15\% (3/20) & \textbf{95\% (19/20)} & 66.8\%\\ \hline
			\end{tabular}
		}
		\vspace{-10pt}
	\end{table*}

	Next, we evaluated the learned models on the robot. 
	In these experiments, we had the robot grasp a given object after executing a series of regrasp actions. 
	Each grasp begins by randomly sampling an end-effector position and angle with the manually engineered system used for the data collection of \sec{sec:setting}, but without closing the fingers of the robot.
	Since we start from a configuration where the fingers are not in contact, it is impossible to fairly compare against the tactile-only variant of our model, which requires the robot to already be in contact with the object to select a meaningful action. 
	Consequently, we compare with the vision-only variant of our model, which is similar to that in~\citep{Levine2016}.
	We then use the learned models to select the next grasp, by solving the optimization of \eq{eq:max-over-next-reward}. 
	For the action optimization, we consider translations in the interval $\left[-2, +2\right]$ \si{\centi\meter}, gripper rotations from $\left[\ang{-17},\ang{+17}\right]$, and force values in $\left[4,25\right]$ \si{\newton}.
	The optimization is performed by randomly sampling 4900 actions, plus 100 additional actions sweeping over the grasping force interval, but having the end-effector rotation and translation set to 0.
	Each action results in performing a translation and rotation of the end-effector, and in closing the fingers with the desired force.
	Moreover, if the predicted grasp success probability is above the desired threshold, the re-grasp also includes lifting the object. 
	In our experiments, we set this threshold to $0.9$.
	To ensure that the probabilities are well-calibrated, we applied Platt scaling~\cite{platt1999probabilistic} to its probability predictions, using a validation set containing approximately $1900$ examples.
    
	As a baseline, we also evaluated against an approach that fits a cylinder around the object using depth data and subsequently attempt to grasp the centroid of the object using a constant grasping force of $10\si{\newton}$. 
	Since we used this cylinder fitting approach as a component of our data collection procedure, it was manually engineered to perform well.

	We first trained the models on 18,070 data points collected as described in \sec{sec:setting}, and evaluated them on a test set of 11 previously unseen objects (that we call ``Easy'').
	These objects significantly differed from the ones seen in the training set in terms of color, weight, shape, friction, etc.  
	From the evaluations, we found that our visuo-tactile model significantly outperformed both the vision-only and the cylinder fitting models, achieving $94\%$ accuracy.
	However, on the harder objects from the ``Hard'' test set, this learned model would not perform very well.
	Hence, we decided to collect more data on the training objects, but this time \emph{on-policy} using the learned model.
	We thus collected a new dataset consisting of 25,404 datapoints, which we used to re-train both the Vision and Tactile+Vision models.
	After retraining, we evaluated the performance again on the ``Hard'' test set.
	In \tab{tab:objects}, we can see how the visuo-tactile model again outperform the other two models.
	Based on these experiments, the largest improvements in performance of our model seem to happen in the presence of compliant objects, and objects where it is difficult to visually ascertain a good grasp, such as small or irregular objects.
        Another interesting result is that the vision-only model performs quite poorly. We hypothesize that the main cause is the relatively small size of the dataset. 
        Prior work~\cite{Levine2016} used a smaller model and 40x more data. 
        As such, it is likely that the performance of our tactile+vision model could also be further improved by collecting more data.

\subsection{Understanding the Learned Visuo-Tactile Model}
\label{sec:results:insideModel}
Our approach relies on a future grasp evaluation model learned entirely from data, without manual specification of heuristically useful behaviors. 
We now examine qualitatively: what strategies has our model learned and what behaviors does it produce?
	
	\subsubsection{Grasping Force}
	\label{sec:results:insideModel:force}
		\begin{figure}[t]
			\centering
			\begin{subfigure}{0.49\linewidth}
			 \centering
			\includegraphics[width=0.99\linewidth]{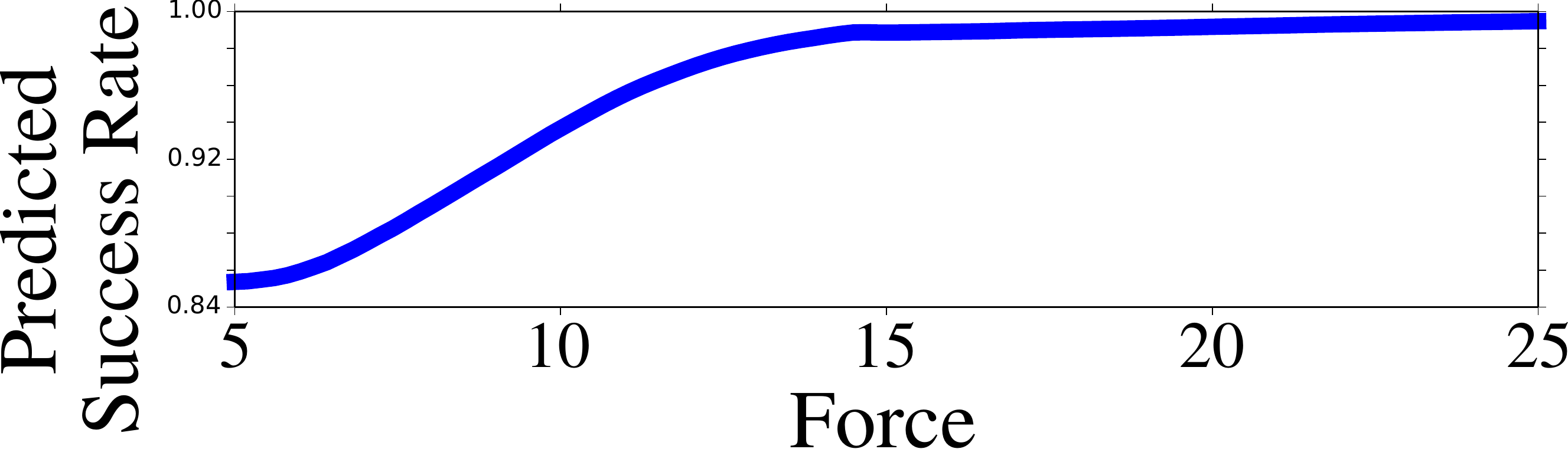}
			\caption{Stable grasp}
			\end{subfigure}
			\hfill
			\begin{subfigure}{0.49\linewidth}
			\includegraphics[width=0.99\linewidth]{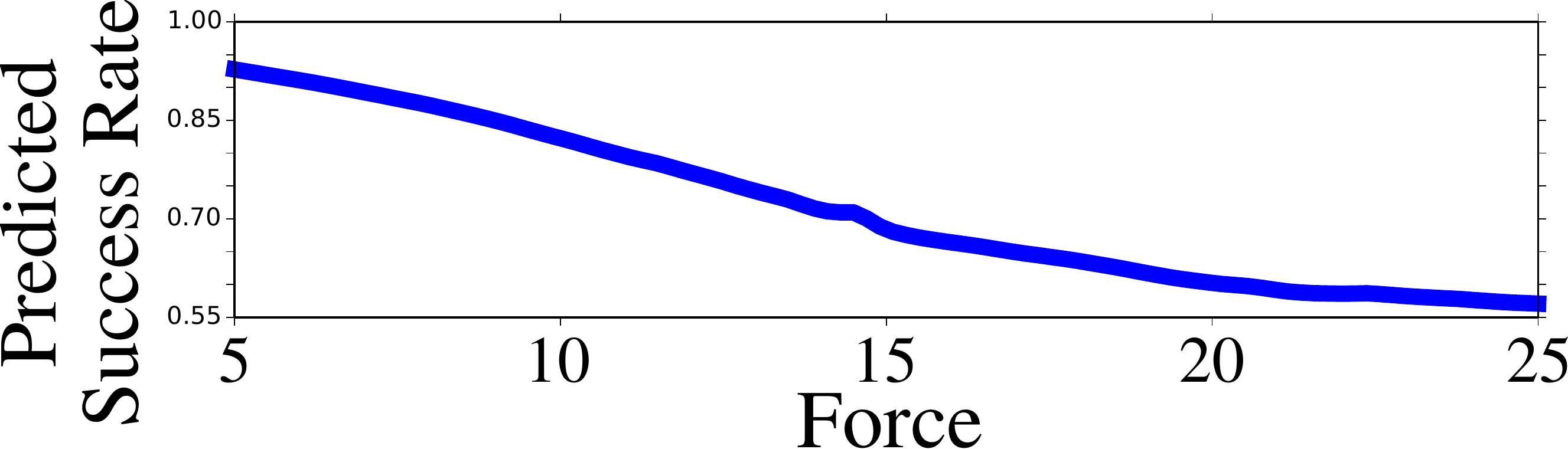}
			\caption{Unstable grasp}
			\label{fig:force_swipe:b}
			\end{subfigure}
			\caption{Predicted grasp success rate with varying the amount of force~$\forcegripper$. The model learned that, when stably in contact with the object, there is a correlation between force applied and success rate. However, for unstable grasps, the model learned that increasing the grasp force might misplace the object and result in an unsuccessful grasp.}
			\label{fig:force_swipe}
			            \vspace{-15pt}
		\end{figure}
		The first question we study is whether or not the model has learned the importance of modulating the amount of force~$\forcegripper$ applied at the fingers for the grasp outcome. 
		Naturally, a stronger grasp is typically more likely to succeed.
		To test this hypothesis, we placed the gripper in a state where it was in contact with a previously unseen object.
		We then asked the model to predict the probability of grasp success given various finger forces, keeping the other parts of the action vector fixed.
		Given this state and candidate actions, we computed the corresponding success rate prediction.
		As illustrated in \fig{fig:force_swipe}, the model appears to have learned that there is a correlation between the force and the grasp outcome.
		However, further analysis shows that the model did not just learn to increase the force in all cases: for multiple situations having very high forces seems to reduce the predicted success rate. 
		For example, we saw this occur when the robot grasped a cube whose corner was only half in contact with the fingers.
		Due to the shape of the fingers, applying large forces in this case would cause the object to be displaced and slip out of the fingers, and the model correctly predicts that lower forces should be preferred (see \fig{fig:force_swipe:b}).
		
	\subsubsection{Height and Center-of-Mass}

		\begin{figure*}
			\begin{minipage}{0.65\linewidth}
				\centering
				\includegraphics[width=0.92\linewidth]{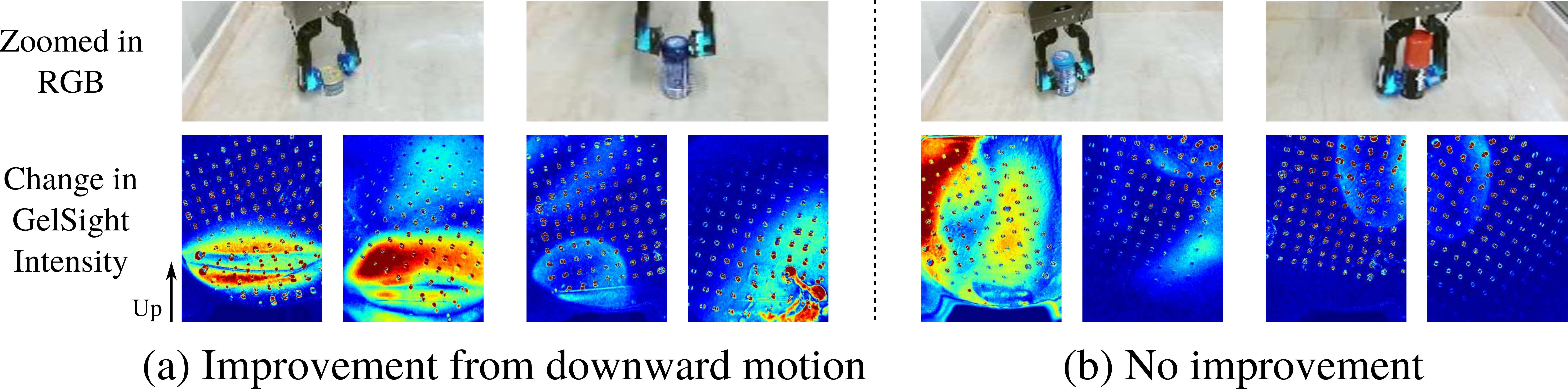}
				\caption{What does the model learn? 
				Here we show examples where the network predicts that a downward motion will result in a grasp with (a) higher or (b) lower chance of succeeding.  
				Notice that downward movement is predicted to be beneficial for cases where the fingers hold the top of an object, but not when they hold it by the bottom.
				To more clearly visualize the contact on the robot's fingertip, we show the change in intensity of the GelSight images.} 
				\label{fig:downward} 
			\end{minipage}
			\hfill
			\begin{minipage}{0.32\linewidth}
				\centering
				\begin{subfigure}{0.49\linewidth}
				  \centering
				\includegraphics[width=\linewidth]{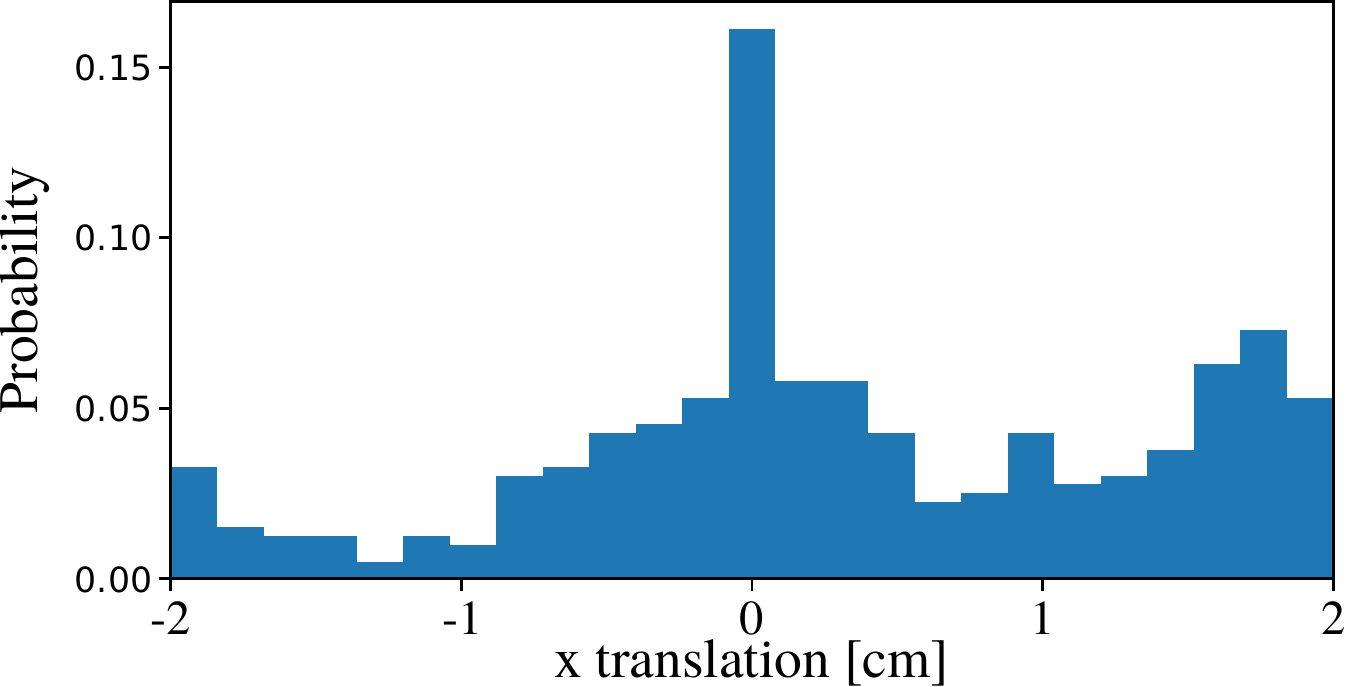}
				\end{subfigure}
				\hfill
				\begin{subfigure}{0.49\linewidth}
				\includegraphics[width=\linewidth]{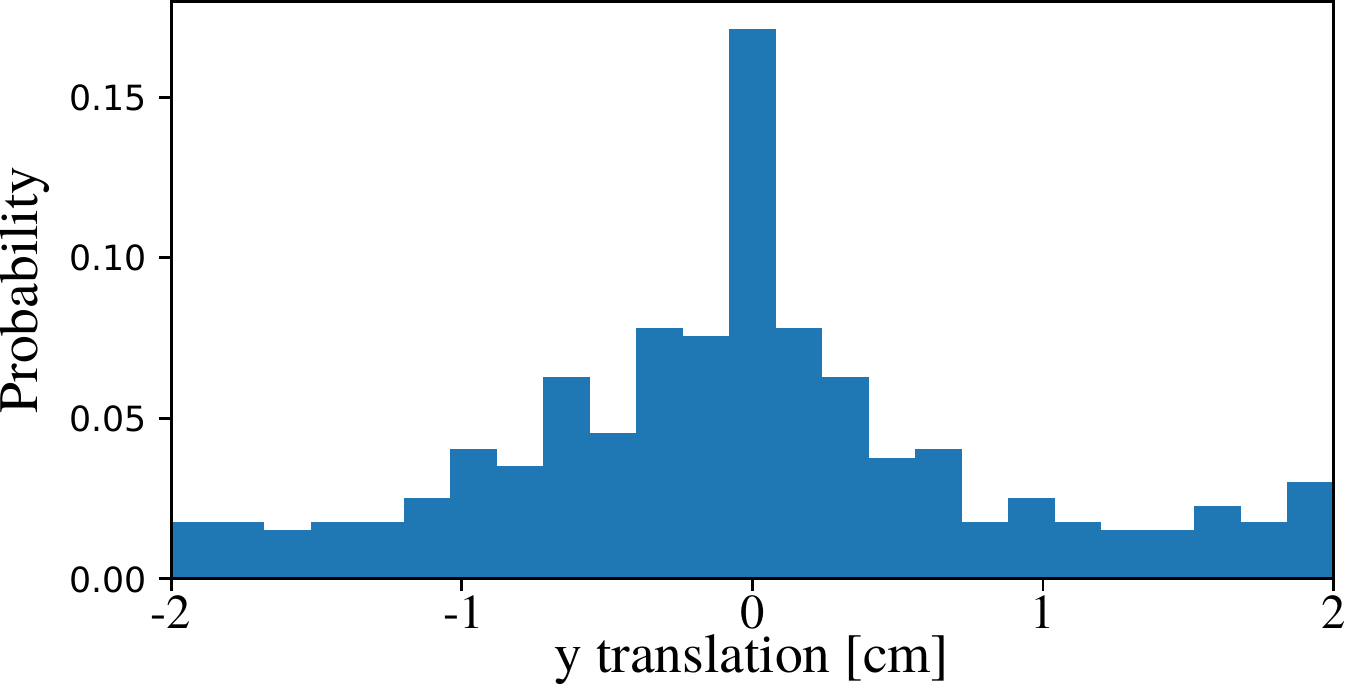}
				\end{subfigure}
				\\\vspace{2pt}
				\begin{subfigure}{0.49\linewidth}
				  \centering
				\includegraphics[width=\linewidth]{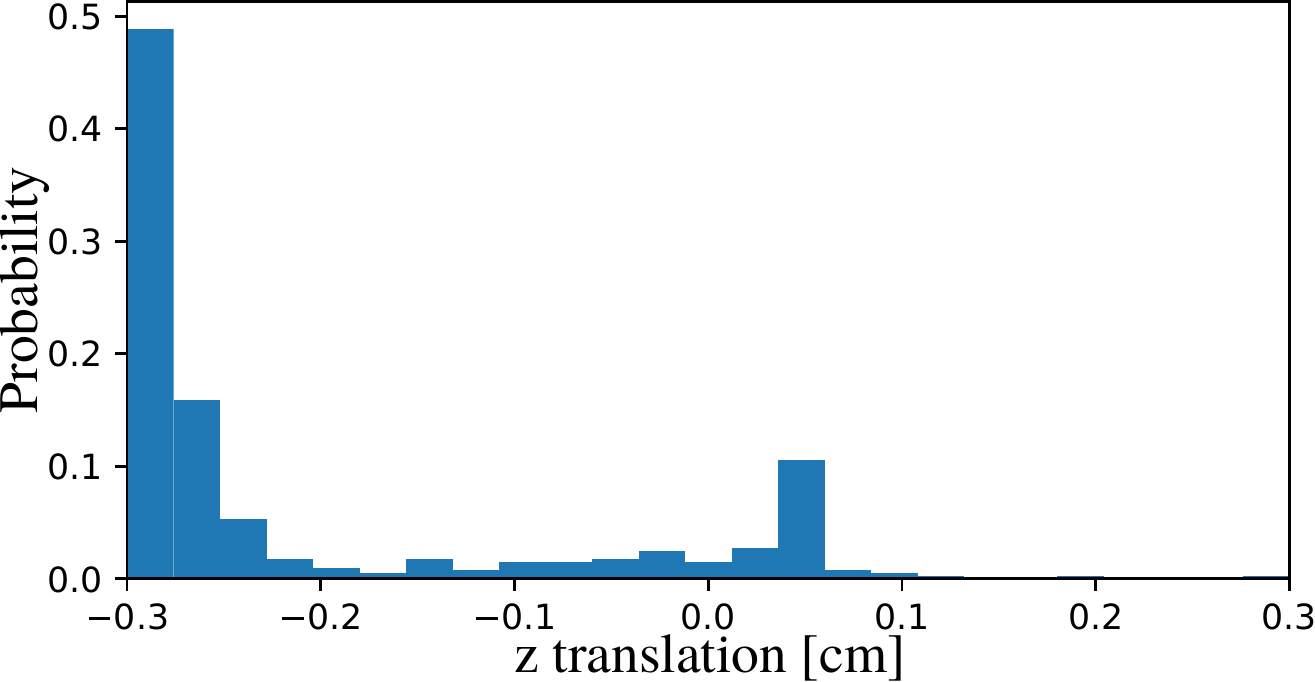}
				\end{subfigure}
				\hfill
				\begin{subfigure}{0.49\linewidth}
				\includegraphics[width=\linewidth]{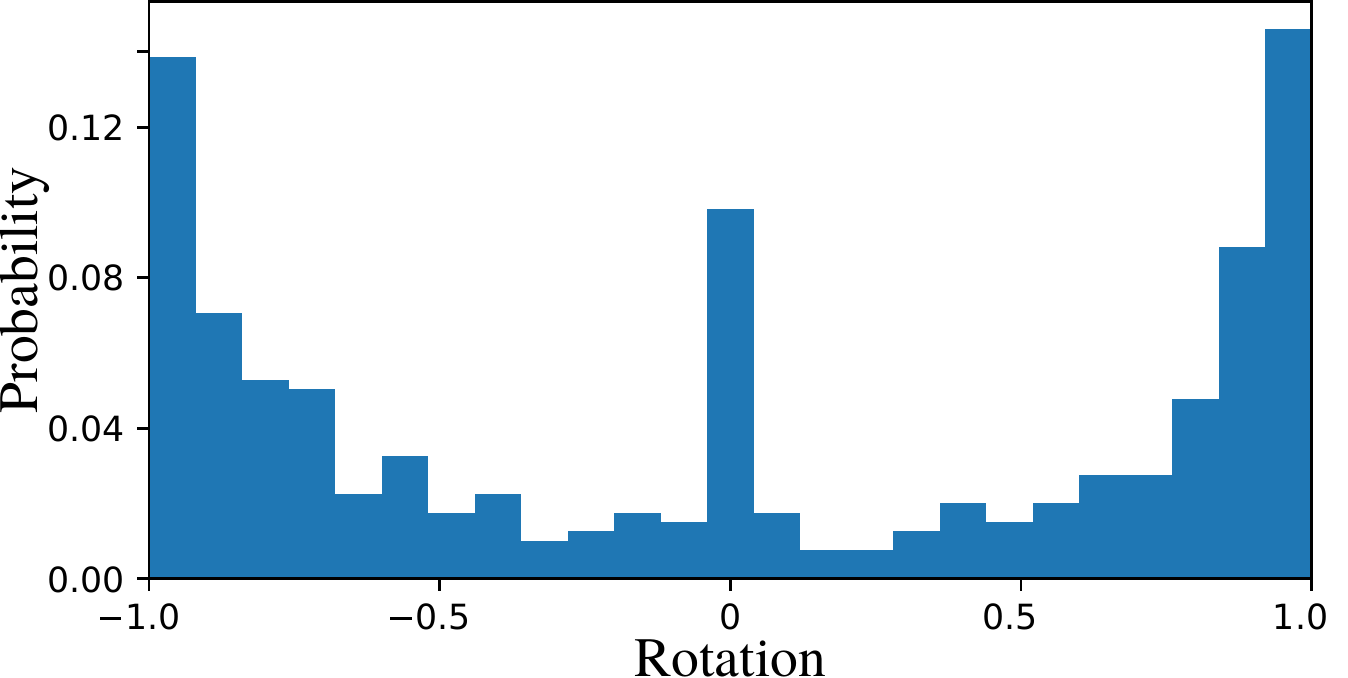}
				\end{subfigure}
				\caption{Histograms of the actions applied by the Tactile+Vision policy for the successful grasps. It can be noticed how the policy strongly favour moving downward.}
				\label{fig:action_hist} 
			\end{minipage}
			\vspace{-10pt}
		\end{figure*}
		\begin{wrapfigure}{r}{0.60\linewidth}
			\centering
			\includegraphics[width=0.98\linewidth]{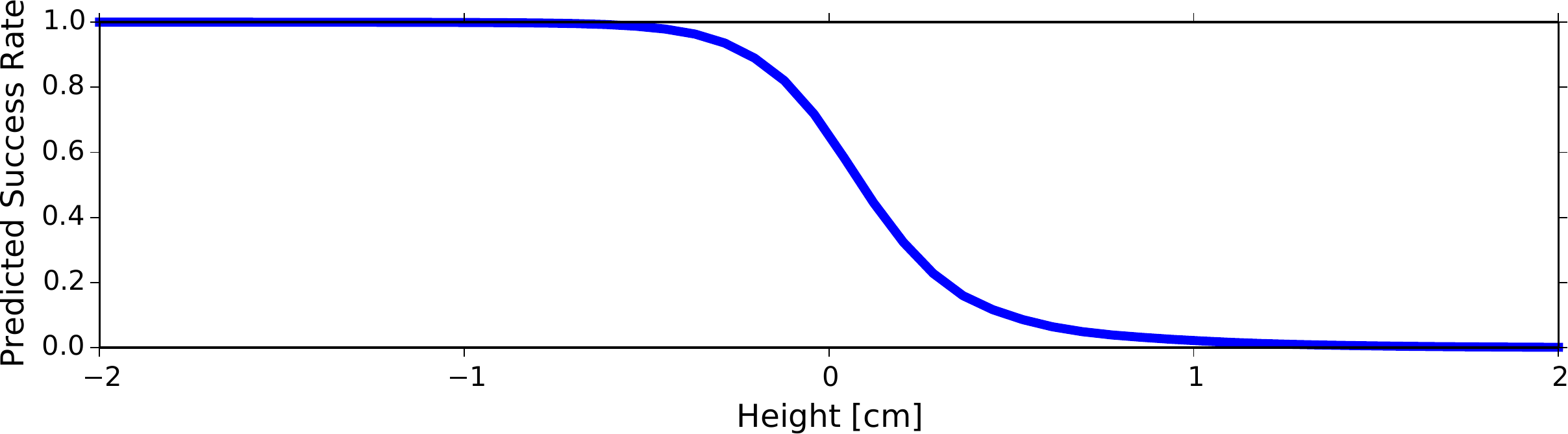}
			\caption{Example of predicted grasp success rate varying the height of the fingers. The model learned that decreasing the height of the fingers generally increases the success rate.}
			\label{fig:height_swipe}
% 			\vspace{-10pt}
		\end{wrapfigure}
		A second important question is what the model learned with respect to the height of the grasp.
		For instance, it may be important to grasp close to the vertical center-of-mass of the object: objects that are held close to their top might slip away under even small perturbations.
		At the same time, objects that are grasped below the center-of-mass might be unstable and rotate around the contact, increasing the chance of slippage.
		Evaluating the model in different circumstances shows that the model learned that the probability of success increases when decreasing the height of the fingers (an example is shown in \fig{fig:height_swipe}).
		The model did not however, seem to have learned any relevant correlation between the height of the object, or the center-of-mass, and the preference for moving downward. 
		In \fig{fig:downward}, we show examples, taken from our dataset, of cases in which the model strongly preferred a downward motion to a static or upward one.  
		For this, we trained a variation of our model without the end effector pose, so that it cannot use the height above the table as a cue. 
		We show held-out examples with the most (and least) predicted improvement in grasp success.  
		The examples with the largest improvement in downward motion tend to be cases in which the top of the object has been gripped (which result in a visible bump in the bottom of the \gelsight{} image).
		\fig{fig:action_hist} shows histograms of the actions performed by the Tactile+Vision model for the successful grasps in \sec{sec:results:grasp}. 
		For the z-translation, almost $50\%$ of the actions used the maximum downward motion allowed (\ie, $2\si{\centi\meter}$), which clearly shows that the learned model acquired a strong preference for moving downward to produce stable grasps.

\subsection{Minimum Force Grasp}
\label{sec:results:minimumForce}
	\begin{figure*}[t]
		\centering
		\begin{subfigure}{0.48\linewidth}
			\centering
			\includegraphics[width=0.84\linewidth]{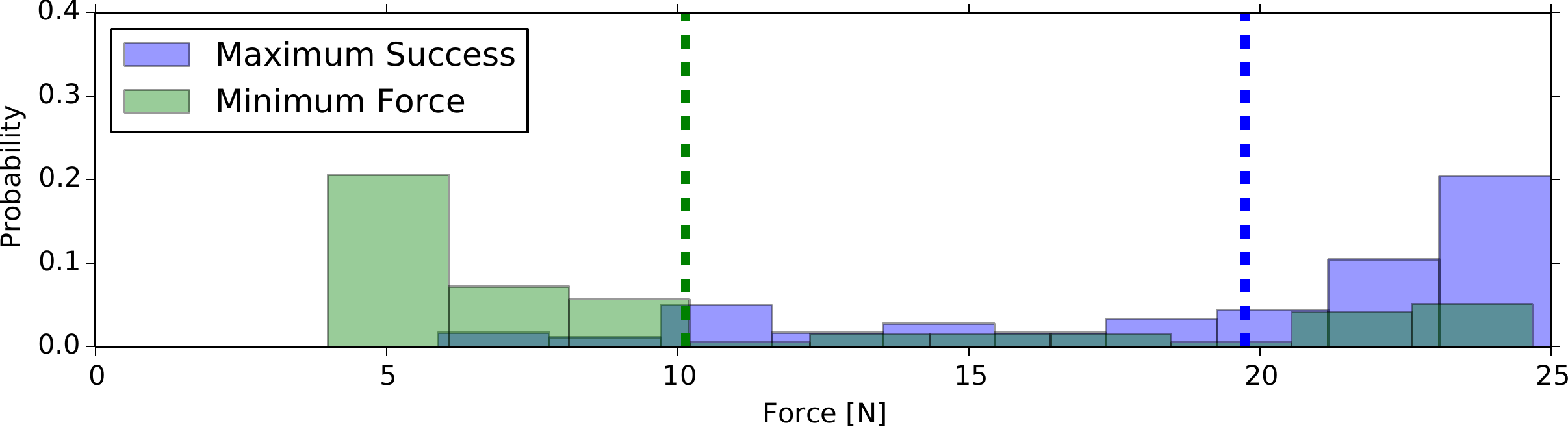}
			\caption{Tactile+Vision}
			\label{fig:force_hist:1}
		\end{subfigure}
		\hfill
		\begin{subfigure}{0.48\linewidth}
			\centering
			\includegraphics[width=0.84\linewidth]{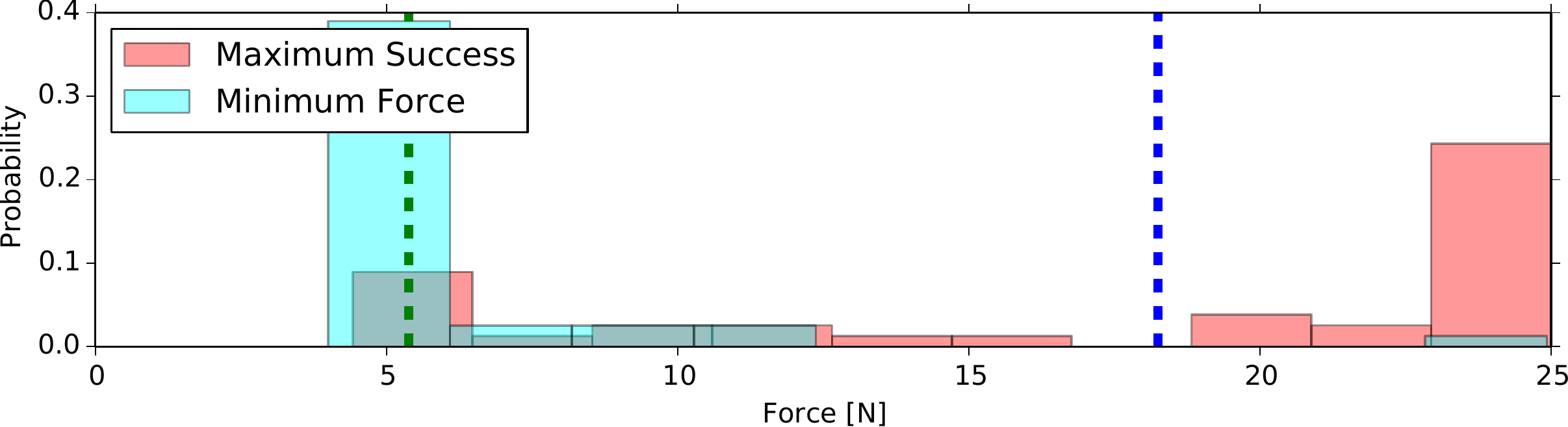}
			\caption{Vision only}
			\label{fig:force_hist:2}
		\end{subfigure}
		\caption{Histogram and mean (dashed lines) of the forces applied in the successful grasps. (a) Although the success rates for the two Tactile+Vision policies are similar ($95\%$ maximum success vs $94\%$ minimum force), the mean force applied is significantly reduced when using the minimum force policy ($10$ vs $20$ \si{\newton}). (b) The success rates for the Vision only policies is lower at $76\%$, but again the mean force applied is significantly reduced when using the minimum force policy ($6$~vs~$18$~\si{\newton}).}
		\label{fig:force_hist}
        \vspace{-15pt}
	\end{figure*}

	One of the benefits of training an action-conditional grasp outcome prediction model, in contrast to the static grasp classification model in prior work~\cite{Calandra2017}, is that we can predict how successful a given grasp will be if we modify the strength of the grasp. 
	Humans typically do not use the strongest grasp possible, but rather employ the minimum amount of contact force, out of consideration for energy consumption and object fragility. 
	Our model also allows us to directly optimize for grasps with either a constraint on the contact force, or via a weighted combination of contact force and grasp success probability.
	In this experiment, we modified the optimization in \eq{eq:max-over-next-reward} as a constrained optimization problem such that the selected action would instead minimize the use of force, but while still having an expected success rate $>90\%$ (if such an action existed, otherwise it would revert to the standard optimization task).

	We evaluated the success rate and applied the force of grasps optimized for either pure grasp success or the minimum force objective on the `Green tea cup' object.
	After evaluating $100$ grasps for each criterion using the Tactile+Vision model, we observed a fairly similar grasp success rate, with $95/100$ successful grasp for the maximum success optimization and $94/100$ for the minimum force grasps.
	However, we can see in \fig{fig:force_hist:1} that, for the successful grasps, the force distribution of the minimum force grasp optimization was substantially lower compared to the maximum success criterion (mean of $10$ vs $20$ \si{\newton}).
	Similar results were obtained also when evaluating the Vision only model, as shown in \fig{fig:force_hist:2}.
	This time, both criteria achieved a success rate of $76\%$ (out of 50 trials), which is lower than the Tactile+Vision model.
	However, the force distribution of the minimum force grasping policy was substantially lower compared to the maximum success criteria at $6$ vs $18$ \si{\newton}.
	These results suggest that using a minimum force optimization with our learned model can effectively reduce the amount of force exerted when grasping, without impacting performance. 
	We believe that this is an important result that show the quality of the learned visuo-tactile model, and further motivate the use of tactile sensors in applications which require handling of fragile objects (\ie, glass or fruit, such as strawberries).

%% file: 6_conclusion.tex
% Main issue
Touch sensing is an inherently active sensing modality, and it is natural that it would be best used in an active fashion, via feedback controllers that incorporate tactile inputs \emph{during} the grasping process. Designing such controllers is challenging, particularly with complex, high-bandwidth tactile sensing combined with visual inputs.
In this paper, we introduced a novel action-conditional deep model capable of incorporating raw inputs from vision and touch.
By using raw visuo-tactile information, this model can continuously re-plan what action to take so as to best grasp objects.
To train this model, we collected over \approxndatagraspingreal{} trials from \nTrainingObj{} training objects.
The learned model is capable of grasping a wide range of unseen objects, and with a high success rate.
Moreover, we demonstrated that with an action-conditioned model, we can easily decrease the amount of force exerted when grasping, while preserving a similar chance of success.  

% Future work
Our method has multiple limitations that could be addressed in future work. 
First, our action-conditioned model only makes single-step predictions, and does not perform information-gathering actions. 
Second, we consider relatively coarse actions -- A model using fine-grained actions could more delicately manipulate the object before the grasp, and potentially react to slippage during the lift-off.
Finally, it would be valuable to extend our approach to more realistic cluttered environments.
Together, addressing these limitations would require a transition to more continuous feedback control strategy (potentially using torque control), which is an exciting avenue for future work.